\documentclass[9pt,conference,a4paper]{IEEEtran}
\IEEEoverridecommandlockouts
\usepackage{cite}
\usepackage{amsmath,amssymb,amsfonts}
\usepackage{algorithmic}
\usepackage{graphicx}
\usepackage{epstopdf}
\usepackage{textcomp}
\usepackage{xcolor}
\def\BibTeX{{\rm B\kern-.05em{\sc i\kern-.025em b}\kern-.08em
    T\kern-.1667em\lower.7ex\hbox{E}\kern-.125emX}}
\begin{document}

\title{UHP-SOT: An Unsupervised High-Performance \\
Single Object Tracker}

\author{Zhiruo Zhou$^{1}$, Hongyu Fu$^{1}$, Suya You$^{2}$, Christoph C. Borel-Donohue$^{2}$ and C.-C. Jay Kuo$^{1}$ \\
$^{1}$ Media Communications Lab, University of Southern California, Los Angeles, CA, USA \\ 
$^{2}$ Army Research Laboratory, Adelphi, Maryland, USA}

\maketitle

\begin{abstract}

An unsupervised online object tracking method that exploits both
foreground and background correlations is proposed and named UHP-SOT
(Unsupervised High-Performance Single Object Tracker) in this work.
UHP-SOT consists of three modules: 1) appearance model update, 2)
background motion modeling, and 3) trajectory-based box
prediction.  A state-of-the-art discriminative correlation filters (DCF)
based tracker is adopted by UHP-SOT as the first module. We point out
shortcomings of using the first module alone such as failure in
recovering from tracking loss and inflexibility in object box
adaptation and then propose the second and third modules to overcome
them.  Both are novel in single object tracking (SOT). We test UHP-SOT on
two popular object tracking benchmarks, TB-50 and TB-100, and show that
it outperforms all previous unsupervised SOT methods, achieves a
performance comparable with the best supervised deep-learning-based SOT
methods, and operates at a fast speed (i.e.  22.7-32.0 FPS on a CPU). 

\end{abstract}


\section{Introduction}\label{sec:introduction}

Video object tracking is one of the fundamental computer vision problems
and has found rich applications in video surveillance
\cite{xing2010multiple}, autonomous navigation \cite{janai2020computer},
robotics vision \cite{zhang2015good}, etc. In the setting of online
single object tracking (SOT), a tracker is given a bounding box on the
target object at the first frame and then predicts its boxes
for all remaining frames \cite{yilmaz2006object}.  Online tracking
methods can be categorized into two categories, unsupervised and supervised \cite{fiaz2019handcrafted}. 
Traditional trackers are
unsupervised.  Recent deep-learning-based (DL-based) trackers demand supervision.
Unsupervised trackers are attractive since they do not need annotated
boxes to train supervised trackers. The performance of trackers
can be measured in terms of accuracy (higher success rate), robustness
(automatic recovery from tracking loss), and speed (higher FPS). 

We examine the design of an unsupervised high-performance tracker and
name it UHP-SOT (Unsupervised High-Performance Single Object Tracker) in
this work.  UHP-SOT consists of three modules: 1) appearance model update,
2) background motion modeling, and 3) trajectory-based box
prediction. Previous unsupervised trackers pay attention to efficient
and effective appearance model update.  Built upon this foundation, an
unsupervised discriminative-correlation-filters-based (DCF-based) tracker
is adopted by UHP-SOT as the baseline in the first module. Yet, the use of
the first module alone has shortcomings such as failure in tracking loss
recovery and being weak in box size adaptation.  We propose
ideas for background motion modeling and trajectory-based box
prediction. Both are novel in SOT.  We test UHP-SOT on two popular object
tracking benchmarks, TB-50 and TB-100 \cite{7001050}, and show that it
outperforms all previous unsupervised SOT methods, achieves a
performance comparable with the best supervised DL-based SOT methods, 
and operates at a fast speed (22.7-32.0 FPS on a CPU). 


\section{Related Work}\label{sec:review}

{\bf DCF-based trackers.} DCF-based trackers provide an efficient
unsupervised SOT solution with quite a few variants, e.g.,
\cite{bolme2010visual, henriques2014high, danelljan2016discriminative,
danelljan2016beyond, bertinetto2016staple, li2018learning,
danelljan2015convolutional, valmadre2017end}. Generally speaking, they
conduct dense sampling around the object patch and solve a rigid
regression problem to learn a template for similarity matching. Under
the periodic assumption of dense samples, they learn the template
efficiently in the Fourier domain and achieve a fast tracking speed.
Spatial-temporal regularized correlation filters (STRCF) \cite{li2018learning} adds spatial-temporal regularization to the
template learning process and performs favorably against other methods
\cite{li2019siamrpn++, danelljan2017eco, hong2015multi,
held2016learning}. 

{\bf DL-based trackers.} DL-based trackers offer a supervised SOT
solution. Some of them use a pre-trained network
\cite{krizhevsky2012imagenet,chatfield2014return} as a feature extractor
and do online tracking based on extracted deep features
\cite{danelljan2017eco,danelljan2016beyond,
ma2015hierarchical,qi2016hedged,wang2018multi}.  Others adopt an
end-to-end model that is either trained by offline video datasets
\cite{li2018high, li2019siamrpn++} or adapted to the online video frames
\cite{lu2018deep,nam2016learning,pu2018deep,song2017crest}.  Recently,
Siamese trackers \cite{bertinetto2016fully, li2018high, li2019siamrpn++,
tao2016siamese,zhu2018distractor,wang2018learning,he2018twofold} are
gaining attention due to the simplicity and effectiveness. They treat
the tracking as a problem of template matching in a large search region.
Unsupervised training in large-scale offline datasets was investigated
in \cite{wang2013learning} and \cite{wang2019unsupervised}.

\section{Proposed UHP-SOT Method}\label{sec:method}

\begin{figure*}[htbp]
\centerline{\includegraphics[width=\textwidth]{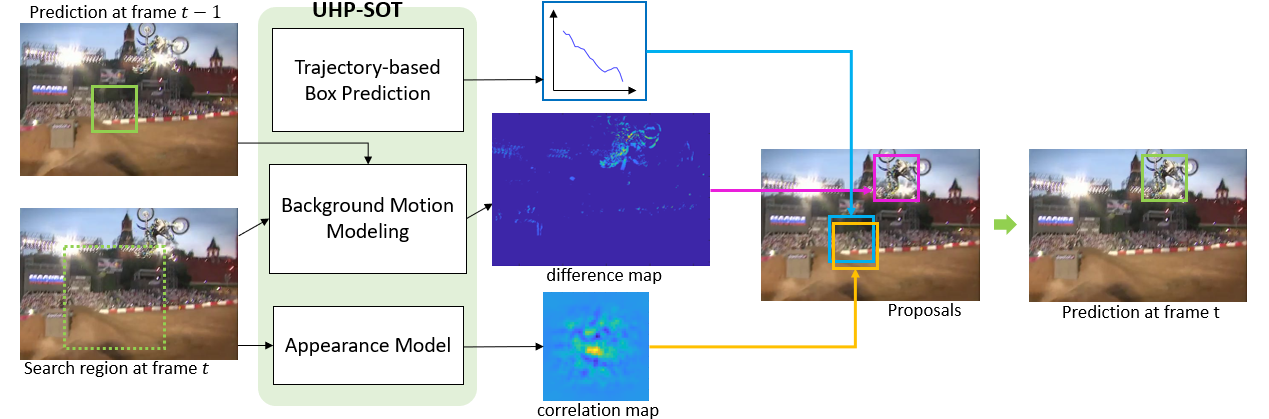}}
\caption{The system diagram of the proposed UHP-SOT method. In the example, the object was lost at time $t-1$ but gets retrieved at time $t$ because the proposal from background motion modeling is accepted.} \label{fig:system}
\end{figure*}

{\bf System Overview.} There are three main challenges in SOT: 
\begin{enumerate}
\item significant change of object appearance, 
\item loss of tracking,
\item rapid variation of object's location and/or shape.
\end{enumerate}
To address these challenges, we propose a new tracker, UHP-SOT, whose
system diagram is shown in Fig. \ref{fig:system}. As shown in the
figure, it consists of three modules:
\begin{enumerate}
\item appearance model update, 
\item background motion modeling,
\item trajectory-based box prediction.
\end{enumerate}

\begin{figure}[htbp]
\centerline{\includegraphics[width=\linewidth]{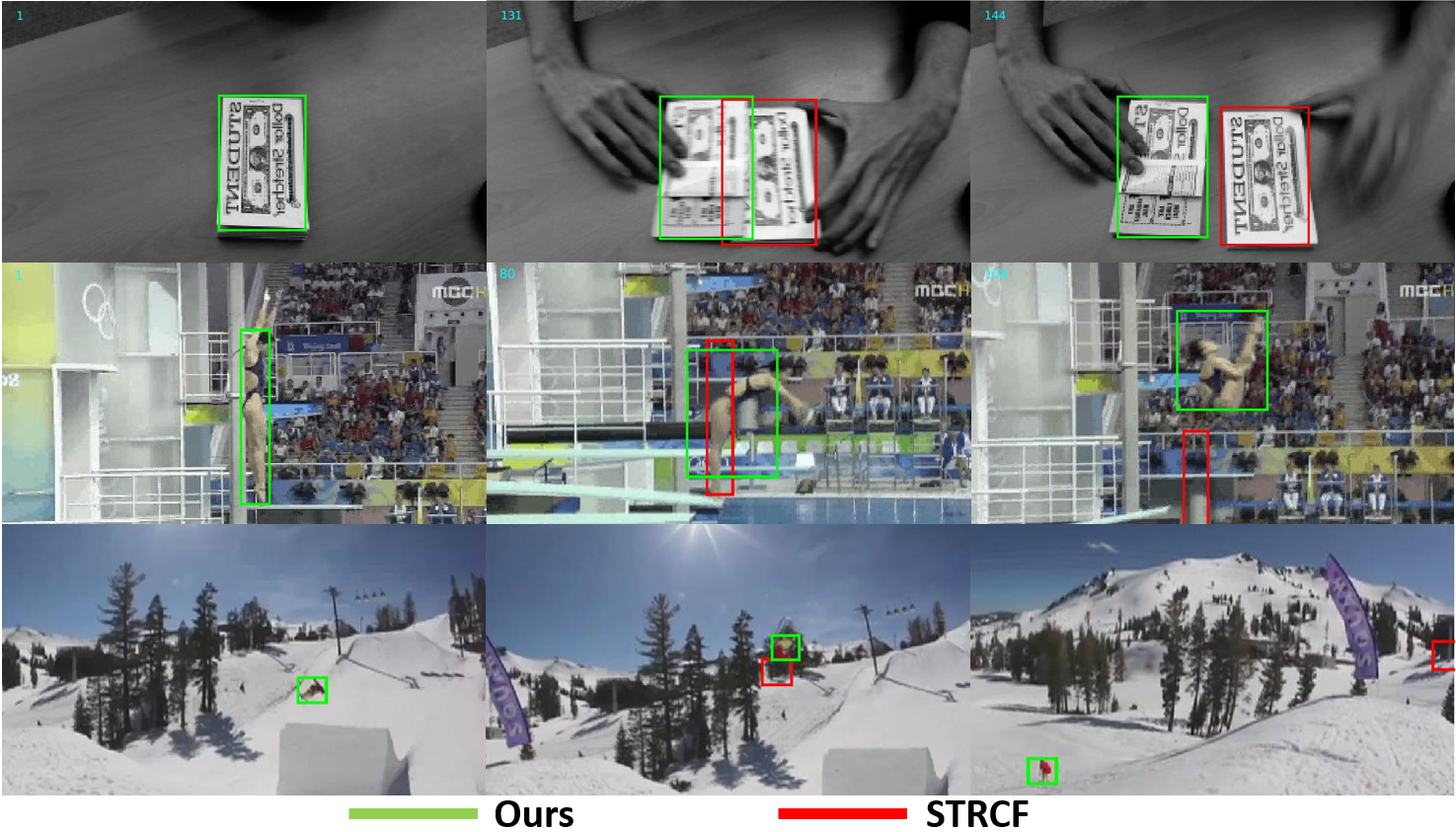}}
\caption{Comparison of the tracking performance of STRCF (red) and 
UHP-SOT (green), where the results of UHP-SOT are closer to the ground truth and those of STRCF drift away.}
\label{fig:showcase}
\end{figure}

UHP-SOT follows the classic tracking-by-detection paradigm where the
object is detected within a region centered at its last predicted
location at each frame.  The histogram of gradients (HOG) and color name
(CN) \cite{danelljan2014adaptive} features are extracted to yield the
feature map. We choose the STRCF tracker \cite{li2018learning} as the
baseline tracker. However, STRCF cannot handle the second and the third
challenges well, as shown in Fig. \ref{fig:showcase}. We propose the
second and the third modules in UHP-SOT to address them. They are the main
contributions of this work.  UHP-SOT operates in the following fashion.
The baseline tracker gets initialized at the first frame. For the
following frames, UHP-SOT gets proposals from all three modules and
chooses one of them as the final prediction based on a fusion strategy.
They are elaborated below. 

{\bf Spatial-temporal regularized correlation filters (STRCF).} In STRCF, the object appearance at frame $t$ is modeled by
a template denoted by $\mathbf{f}_t$. It is used for similarity matching
at frame $(t+1)$.  The template is initialized at the first frame.  Assume
that the cropped patch centered at the object location has a size of
$N_x \times N_y$ pixels at frame $t$.  Then, the template gets updated at
frame $t$ by solving the following regression equation:
\begin{equation}
\mathrm{arg}\min_{\mathbf{f}} \frac{1}{2}\|\sum_{d=1}^D \mathbf{x}_{t}^{d}*
\mathbf{f}^{d}-\mathbf{y}\|^2 + \frac{1}{2}\sum_{d=1}^D \|\mathbf{w}\cdot 
\mathbf{f}^{d} \|^2 + \frac{\mu}{2}\|\mathbf{f}-\mathbf{f}_{t-1} \|^2,
\nonumber
\end{equation}
where $y \in \mathbb{R}^{N_x \times N_y}$ is a centered Gaussian-shaped
map used as regression labels, $\mathbf{x}_{t} \in \mathbb{R}^{N_x
\times N_y \times D}$ is the spatial map of $D$ features, $*$ denotes
the spatial convolution of the same feature, $\mathbf{w}$ is the spatial
weight on the template, $\mathbf{f}_{t-1}$ is the template obtained from
time $t-1$, and $\mu$ is a constant regularization coefficient.  We can
interpret the three terms in the right-hand-side of the above equation
as follows.  The first term demands that the new template has to match
the newly observed features accordingly with the assigned labels. The
second term is the spatial regularization term which demands that
regions outside of the box contribute less to the matching
result. The third term corresponds to self-regularization that ensures
smooth appearance change. To search for the box in frame
$(t+1)$, STRCF correlates template $\mathbf{f}_t$ with the search region
and determines the box by finding the location that gives the
highest response. Although STRCF can model the appearance change for
most sequences, it suffers from overfitting so that it is not able to
adapt to largely deformable objects quickly. Furthermore, it cannot
recover after tracking loss. 

The template model $\mathbf{f}$ is updated at every frame with a fixed
regularization coefficient $\mu$ in STRCF. In our implementation, we
skip updating $\mathbf{f}$ if no obvious motion is observed. In
addition, a smaller $\mu$ is used when all modules agree with each other
in prediction so that $\mathbf{f}$ can quickly adapt to the new
appearance for largely deformable objects. 

{\bf Background motion modeling.} For SOT, we can decompose the pixel
displacement between adjacent frames ( also called optical flow) into
two types: object motion and background motion.  Background motion is
usually simpler so that it can be well modeled by a parametric model.
Background motion estimation \cite{hariharakrishnan2005fast,
aggarwal2006object} finds applications in video stabilization, coding
and visual tracking. Here, we propose a 6-parameter model in form of
$$
x_{t+1}=\alpha_1 x_t + \alpha_2 y_t + \alpha_0, \mbox{  and  }
y_{t+1}=\beta_1 x_t + \beta_2 y_t + \beta_0, 
$$
where $(x_{t+1},y_{t+1})$ and $(x_t,y_t)$ are corresponding background
points in frames $(t+1)$ and $t$, and $\alpha_i$ and $\beta_i$, $i=0,1,2$
are model parameters. Given more than three pairs of corresponding
points, we can solve the model parameters using the linear least-squares
method. Usually, we choose a few salient points (e.g., corners) to build
the correspondence and determine the parameters. We apply the background
model to the grayscale image $I_t(x,y)$ of frame $t$ to find the estimated $\hat{I}_{t+1}(x,y)$ of
frame $(t+1)$. Afterwards, we can compute the difference map $\Delta I$:
$$
\Delta I = \hat{I}_{t+1}(x,y) - I_{t+1}(x,y)
$$
which is expected to have small and large values in the background and
foreground regions, respectively. Thus, we can determine potential
object locations.  While the DCF-based baseline exploits foreground
correlation to locate the object, background modeling uses background
correlation to eliminate background influence in object tracking. They
complement each other for some challenging task such as recovery from
tracking loss.  The DCF-based tracker cannot recover from tracking loss
easily since it does not have a global view of the scene. In contrast,
our background modeling module can still find potential locations of the
object by removing the background region. 

\begin{figure}[!htbp]
\centerline{\includegraphics[width=0.95\linewidth]{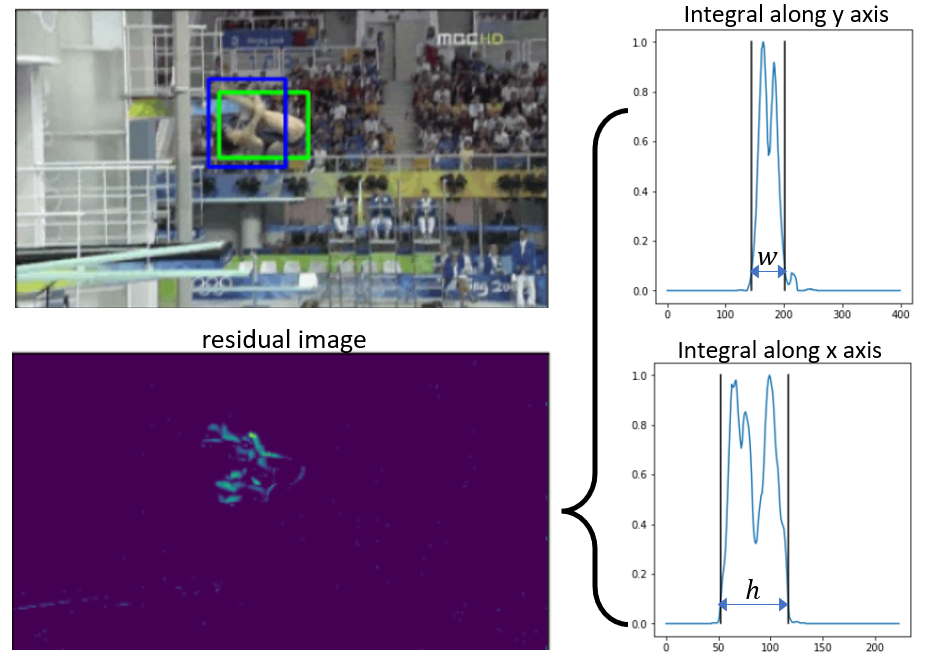}}
\caption{Illustration of shape change estimation based on background
motion model and the trajectory-based box prediction, where
the ground truth and our proposal are in green and blue, respectively.} 
\label{fig:shape}
\end{figure}

{\bf Trajectory-based Box Prediction.} Given box
centers of last $N$ frames, $\{
(x_{t-N},y_{t-N}),\cdots,(x_{t-1},y_{t-1}) \}$, we calculate $N-1$
displacement vectors $\{ (\Delta x_{t-N+1},\Delta y_{t-N+1}),\cdots,(\Delta
x_{t-1},\Delta y_{t-1}) \}$ and apply the principal component analysis
(PCA) to them.  To predict the displacement at frame $t$, we fit the
first principal component using a line and set the second principal
component to zero to remove noise.  Then, the center location of the
box at frame $t$ can be written as
$$
(\hat{x}_{t}, \hat{y}_{t}) = (x_{t-1}, y_{t-1}) + (\hat{\Delta x}_{t}, 
\hat{\Delta y}_{t}).
$$
Similarly, we can estimate the width and the height of the box
at frame $t$, denoted by $(\hat{w}_t, \hat{h}_t)$.  Typically, the
physical motion of an object has an inertia in motion trajectory and its
size, and the box prediction process attempts to maintain the
inertia. It contributes to better tracking performance in two ways.
First, it can remove small fluctuation of the box in its
location and size. Second, when there is a rapid deformation of the
target object, the appearance model alone cannot capture the shape
change effectively. In contrast, the combination of background motion
modeling and the trajectory-based box prediction can offer a
more satisfactory solution. An example is given in Fig. \ref{fig:shape},
which shows a frame of the \textit{diving} sequence in the upper-left subfigure,
where the green and the blue boxes are the ground truth and the
result of UHP-SOT, respectively.  Although a DCF-based tracker can detect
the size change by comparing correlation scores at five image
resolutions, it cannot estimate the change of its aspect ratio.  In
contrast, the residual image after background removal in UHP-SOT, as shown
in the lower-left subfigure, reveals the object shape. We sum up the
absolute pixel values of the residual image horizontally and vertically.
We use a threshold to determine the two ends of an interval. Then,
we have
$$
\hat{w}=x_{\max}-x_{\min}, \mbox{ and } \hat{h}=y_{\max}-y_{\min}. 
$$
Note that the raw estimation may not be stable across different frames.
Estimations that deviate too much from the trajectory of $(\Delta
w_{t},\Delta h_{t})$ are rejected. Then, we have a robust yet flexibly
deformable box proposal. 

{\bf Fusion strategy.} We have three box proposals for the target object
at frame $t$: 1) $B_\mathrm{app}$ from the baseline tracker to capture
appearance change, 2) $B_\mathrm{trj}$ from the trajectory predictor to
maintain the inertia of box position/shape and 3) $B_\mathrm{bgd}$ from
the background motion predictor to eliminate unlikely object regions. We
need a fusion strategy to yield the final box location/shape, which is
described below.  We store two template models: the latest model
$\mathbf{f}_{t-1}$, and an older model, $\mathbf{f}_{i}$, $i\leq t-1$,
where $i$ is the last time step where all three boxes have the same
location.  Model $\mathbf{f}_{i}$ is less likely to be contaminated
since it needs agreement from all modules while $B_\mathrm{bgd}$ can
jump around. To check the reliability of the three proposals, we compute
correlation scores between three pairs: ($\mathbf{f}_{t-1}$,
$B_\mathrm{app}$), ($\mathbf{f}_{t-1}$, $B_\mathrm{trj}$), and
($\mathbf{f}_{i}$, $B_\mathrm{bgd}$) and then apply a rule-based 
fusion strategy:
\begin{itemize}
\item General rule: Choose the proposal with the highest score.
\item Special rule: Although $B_\mathrm{app}$ has the highest score, we
observe that $B_\mathrm{trj}$ has a close score, agrees with
$B_\mathrm{bgd}$, and reveals sudden jump (say, larger than 30 pixels).
Then, we choose $B_\mathrm{trj}$ instead. 
\end{itemize}

\begin{figure*}[htbp]
\centerline{\includegraphics[width=\textwidth]{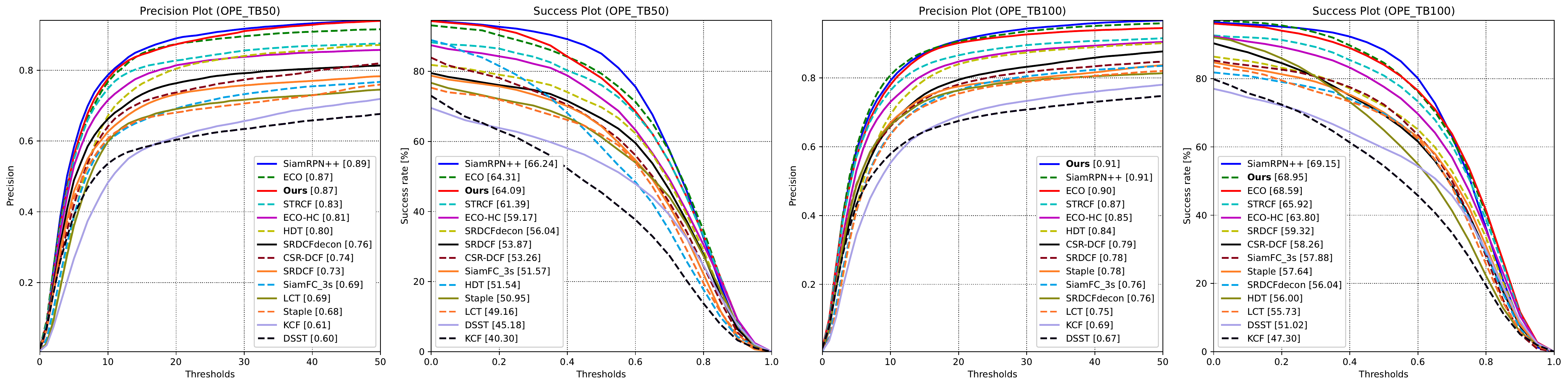}}
\caption{The precision plot and the success plot on TB-50 and TB-100. To rank different methods, the distance precision is measured
at 20-pixel threshold, and the overlap precision
is measured by the AUC score. We collect other trackers’ raw results from the official websites to generate results.}
\label{fig:benchmark}
\end{figure*}

\begin{figure*}[htbp]
\centerline{\includegraphics[width=\textwidth]{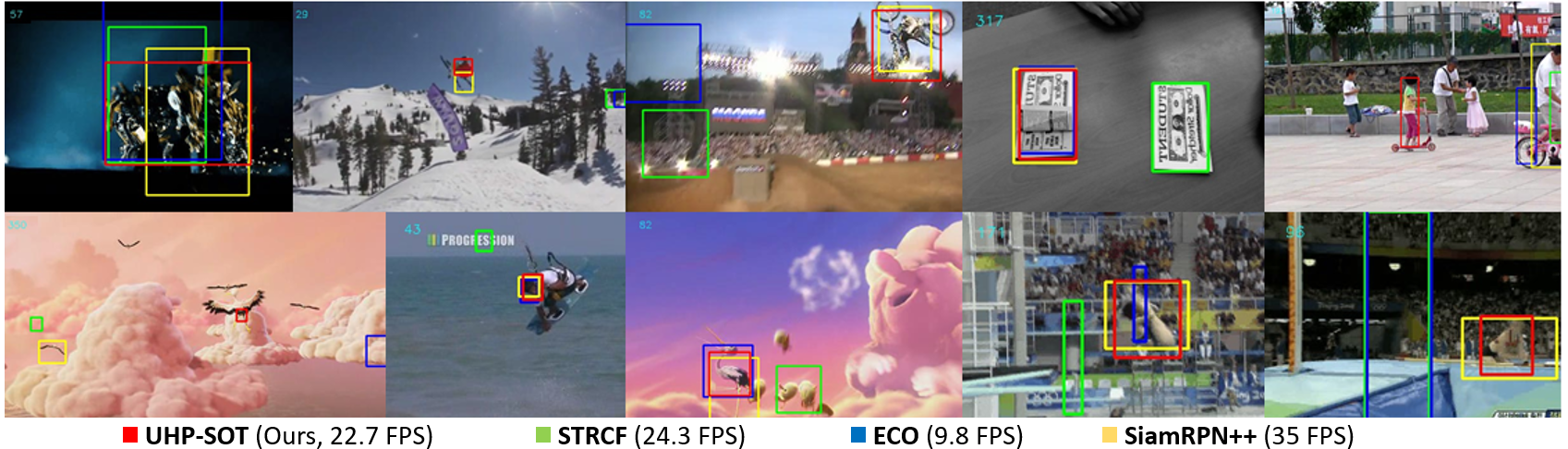}}
\caption{Qualitative evaluation of UHP-SOT, STRCF \cite{li2018learning},
ECO \cite{danelljan2017eco} and SiamRPN++ \cite{li2019siamrpn++}
on 10 challenging videos from TB-100. They are (from left to right and
top to down): \textit{Trans}, \textit{Skiing}, \textit{MotorRolling},
\textit{Coupon}, \textit{Girl2}, \textit{Bird1}, \textit{KiteSurf},
\textit{Bird2}, \textit{Diving}, \textit{Jump}, respectively.}
\label{fig:compare_top4}
\end{figure*}

\section{Experiments}\label{sec:experiments}

{\bf Experimental Set-up.} We compare UHP-SOT with state-of-the-art
unsupervised and supervised trackers on TB-50 and TB-100 datasets
\cite{7001050}. The latter contains $100$ videos and $59,040$ frames.
Evaluation is performed based on the ``One Pass Evaluation (OPE)"
protocol.  Evaluation metrics include the precision plot (i.e., the
distance of the predicted and actual box centers) and the success plot
(i.e., overlapping ratios at various thresholds). The distance precision
is measured at 20-pixel threshold to rank different methods. The overlap
precision is measured by the area-under-curve (AUC) score.  We use the
same hyperparameter settings in STRCF except regularization coefficient
$\mu$.  STRCF sets $\mu=15$ while UHP-SOT selects $\mu \in\{10,5,0\}$ if
the appearance box is not chosen. The smaller the correlation score, the
smaller $\mu$.  We set $N=20$ in the number of previous frames for
trajectory prediction.  The cutting threshold along horizontal/vertical
directions is set to 0.1. UHP-SOT runs at 22.7 FPS on a PC equipped with
an Intel(R) Core(TM) i5-9400F CPU, maintaining a near real-time speed
(while STRCF operates at a speed of 24.3 FPS). 

{\bf Performance Evaluation.} Fig. \ref{fig:benchmark} compares UHP-SOT
with state-of-the-art unsupervised trackers ECO-HC
\cite{danelljan2017eco}, STRCF \cite{li2018learning}, SRDCFdecon
\cite{danelljan2016adaptive}, CSR-DCF \cite{alan2018discriminative},
SRDCF \cite{danelljan2015learning}, Staple \cite{bertinetto2016staple},
KCF \cite{henriques2014high}, DSST \cite{danelljan2016discriminative}
and supervised trackers SiamRPN++ \cite{li2019siamrpn++}, ECO
\cite{danelljan2017eco}, HDT \cite{fiaz2019handcrafted}, SiamFC\_3s
\cite{bertinetto2016fully}, LCT \cite{ma2015long}. UHP-SOT outperforms
STRCF by 4\% in precision and 3.03\% in overlap on TB-100 and 4\% in
precision and 2.7\% in overlap on TB-50, respectively.  As an
unsupervised light-weight tracker, UHP-SOT achieves performance comparable
with state-of-the-art deep trackers such as SiamRPN++ with
ResNet-50\cite{he2016deep} as the backbone. UHP-SOT outperforms another
deep tracker, ECO, which uses a Gaussian Mixture Model to store seen
appearances and runs at around 10 FPS, in both accuracy and speed.  

We show results on 10 challenging sequences on TB-100 for the top-4
trackers in Fig. \ref{fig:compare_top4}. Generally, UHP-SOT can follow
small moving objects or largely deformed objects like human body tightly
even though some of them have occlusions. These are attributed to its
quick recovery from tracking loss via background modeling and stability
via trajectory prediction. On the other hand, UHP-SOT does not perform
well for \textit{Ironman} and \textit{Matrix} because of rapid changes
in both the foreground object and background. They exist in movie
content due to editing in movie post-production. They do not occur in
real-world object tracking. Deep trackers perform well for
\textit{Ironman} and \textit{Matrix} by leveraging supervision.

\begin{figure}[htbp]
\centerline{\includegraphics[width=0.95\linewidth]{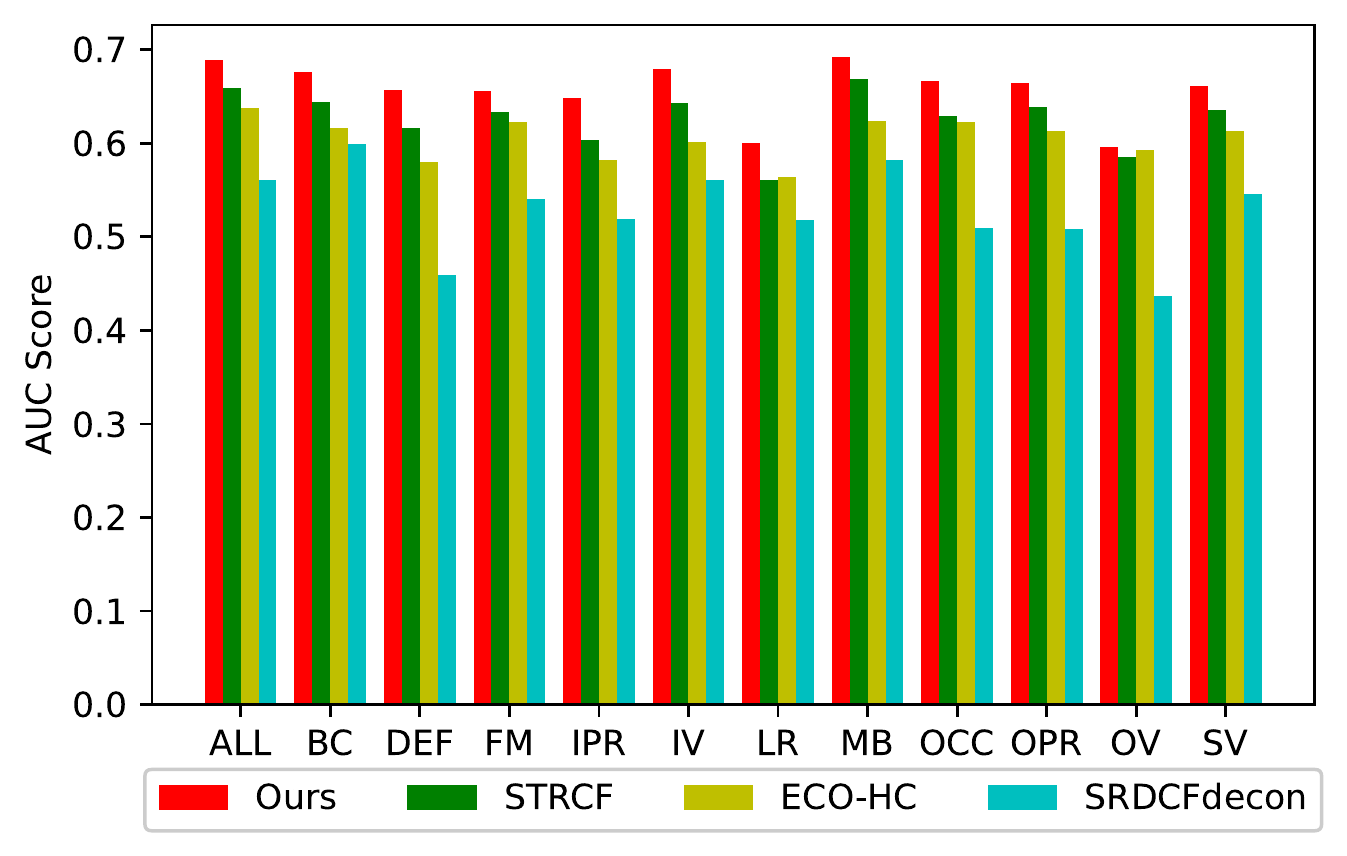}}
\caption{Area-under-curve (AUC) score for attribute-based evaluation on the TB-100 dataset, where the 11
attributes are background clutter (BC), deformation (DEF), fast motion
(FM), in-plane rotation (IPR), illumination variation (IV), low
resolution (LR), motion blur (MB), occlusion (OCC), out-of-plane
rotation (OPR), out-of-view (OV), and scale variation (SV),
respectively.}\label{fig:attrs}
\end{figure}

We further analyze the performance variation under different challenging
tracking scenarios for TB-100. We present the AUC score in Fig.
\ref{fig:attrs} and compare with other state-of-the-art unsupervised DCF
trackers STRCF, ECO-HC, and SRDCFdecon. Our method outperforms other
trackers in all attributes, especially in deformation (DEF), in-plane
rotation (IPR) and low resolution (LR). Difficult sequences in those
attributes include \textit{MotorRolling}, \textit{Jump}, \textit{Diving}
and \textit{Skiing}, where the target appearance changes fast due to large
deformation. It is difficult to reach a high overlapping ratio without
supervision or without an adaptive box aspect ratio strategy. 

Finally, we test two other variants of UHP-SOT: UHP-SOT-I (without
trajectory prediction), and UHP-SOT-II (without background motion
modeling) in Table \ref{tab1:ablation}.  The gap between UHP-SOT and
UHP-SOT-I reveals the importance of inertia provided by trajectory
prediction. UHP-SOT-I shows that background modeling does a good job in
handling some difficult cases that STRCF cannot cope with, leading to a
gain of 2\% in precision and 1.95\% in overlap.  UHP-SOT-II rejects large
trajectory deviation and uses a smaller regularization coefficient to
strengthen this correction.  The accuracy of UHP-SOT-II drops more due to
naive correction without confirmation from background modeling. Yet, it
operates a much faster speed (32.02 FPS). Both background modeling and
trajectory prediction are lightweight modules and run in real time. 

\begin{table}[htbp]
\caption{Performance of UHP-SOT, UHP-SOT-I, UHP-SOT-II and STRCF on the
TB-100 dataset, where AUC is used for success rate.}
\begin{center}
\begin{tabular}{ccccc}
\hline
 & UHP-SOT & UHP-SOT-I & UHP-SOT-II & STRCF\\ \hline
Success (\%) & \textbf{68.95} & 67.87 & 65.64 & 65.92 \\
Precision & \textbf{0.91} & 0.89 & 0.87 & 0.87 \\
Speed (FPS) & 22.73 & 23.68 & \textbf{32.02} & 24.30 \\
\hline
\end{tabular}
\label{tab1:ablation}
\end{center}
\end{table}

\section{Conclusion and Future Work}\label{sec:conclusion}

An unsupervised high-performance tracker called UHP-SOT, which uses STRCF
as the baseline with two novel improvements, was proposed in this work.
They are background motion modeling and a trajectory-based box
prediction. It was shown by experimental results that UHP-SOT offers an
effective real-time tracker in resource-limited platforms.  
Based on our study, object tracking appears to be a low-level vision problem where
supervision may not be critical in all cases. To test this hypothesis we would like to apply supervised
object detectors such as Yolo to edited movie sequences frame by frame 
and demonstrate the merit of supervision in the future.

\clearpage
\bibliographystyle{IEEEtran}
\bibliography{ref}

\end{document}